\PassOptionsToClass{balance=false}{acmart}
\documentclass[sigconf,nonacm]{acmart}
\setcopyright{none}
\usepackage{amsmath,booktabs,graphicx,xurl,placeins,flafter,tabularx}
\usepackage{helvet}
\usepackage{etoolbox,colortbl}
\usepackage{tikz}
\makeatletter
\patchcmd{\hyxmp@parse@acmart}
  {\let\@author=\@gobble}
  {\let\@author=\@gobble\let\@authornotemark\relax\let\@@authornotemark\@gobble}
  {}{}
\makeatother
\definecolor{SHBlue}{HTML}{8D9FCD}
\definecolor{SHGreen}{HTML}{62BE9F}
\definecolor{SHBlueLight}{HTML}{DBE1EC}
\definecolor{SHGreenLight}{HTML}{D0EEE4}
\definecolor{SHInk}{HTML}{263747}
\definecolor{SHGitHub}{HTML}{24292F}
\definecolor{SHLink}{HTML}{586FA0}
\tikzset{SHResourceBadge/.style={rounded corners=4pt,inner xsep=10pt,
  inner ysep=4pt,text height=11pt,text depth=3pt,line width=0.5pt,
  font=\sffamily\fontsize{9}{11}\selectfont}}
\newcommand{\shhomepagebutton}{%
  \href{https://ethanning.github.io/SearchHorizonBench/}{%
    \tikz[baseline=7pt,x=1pt,y=1pt]{%
      \shade[rounded corners=4.5pt,left color=SHLink!82!SHInk,
        right color=SHInk!86!SHGreen] (0,0) rectangle (98,22);
      \draw[rounded corners=4.5pt,SHBlue!65!white,line width=0.45pt]
        (0.2,0.2) rectangle (97.8,21.8);
      \draw[SHBlue!35!white,opacity=0.35,line width=0.4pt,line cap=round]
        (6,20.6)--(92,20.6);
      \begin{scope}[shift={(5,4)}]
        \shade[rounded corners=3.5pt,top color=SHBlue!78!SHLink,
          bottom color=SHLink!85!SHInk] (0,0) rectangle (14,14);
        \draw[rounded corners=3.5pt,SHBlue!55!white,line width=0.35pt]
          (0.2,0.2) rectangle (13.8,13.8);
        \draw[white,line width=0.8pt,line cap=round]
          (3.1,6.2) .. controls (4.0,9.0) and (6.0,10.4) .. (7.9,9.6)
          .. controls (9.4,9.2) and (10.3,7.9) .. (10.9,6.2);
        \draw[SHGreenLight,line width=0.8pt,line cap=round] (3.1,4.5)--(10.9,4.5);
        \fill[SHGreenLight] (8.5,9.4) circle[radius=0.85pt];
      \end{scope}
      \node[anchor=base west,inner sep=0pt,text=white,
        font=\sffamily\fontsize{9}{11}\selectfont] at (24,7) {Homepage};
      \draw[SHGreenLight,line width=0.75pt,line cap=round,line join=round]
        (85,8)--(90,13) (85,13)--(90,13)--(90,8);
    }}}
\newcommand{\shresourcebutton}[3]{\href{#1}{%
  \tikz[baseline=(badge.base)]{\node[SHResourceBadge,#2] (badge) {#3};}}}
\makeatletter
\apptocmd{\@mkauthors}{%
  \global\setbox\mktitle@bx=\vbox{\unvbox\mktitle@bx\unskip\kern3pt
    \hbox to\textwidth{\hfil
      \shhomepagebutton
      \hspace{9pt}%
      \shresourcebutton{https://github.com/EthanNing/SearchHorizonBench}
        {draw=SHGitHub,fill=SHGitHub,text=white}
        {\raisebox{-2.7pt}{\includegraphics[height=11.5pt]{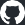}}\hspace{6pt}GitHub}%
      \hfil}}%
}{}{}
\makeatother
\AtBeginEnvironment{tabular}{\sffamily}
\hypersetup{keeppdfinfo}
\newcommand{\SEARCH}{\textsc{Search}}
\newcommand{\THINK}{\textsc{Think}}

\title[SearchHorizonBench]{SearchHorizonBench:\\ A Benchmark for Search Value in AI Agents}
\author{Jingjie Ning}
\authornote{Corresponding author.}
\email{jening@cs.cmu.edu}
\affiliation{\institution{Carnegie Mellon University}\city{Pittsburgh}\state{PA}\country{USA}}
\author{Xueqi Li}
\email{xueqil@cs.cmu.edu}
\affiliation{\institution{Carnegie Mellon University}\city{Pittsburgh}\state{PA}\country{USA}}
\author{Yibo Kong}
\email{yibok@cs.cmu.edu}
\affiliation{\institution{Carnegie Mellon University}\city{Pittsburgh}\state{PA}\country{USA}}

\authorsaddresses{}
\begin{document}
\begin{abstract}
AI agents must decide when another search is worth its cost. Answer-success probabilities capture question difficulty, while this decision requires estimating the additional benefit of search. We introduce SearchHorizonBench, a benchmark for learning and evaluating search value from execution experience. Paired continuations and repeated executions separate expected search gain from generation noise and connect prediction quality to decision utility. The resource contains 45,632 executions across 3,168 questions. On 1,632 held-out questions, historical calibration reduces Brier error by 16.2\%, with the improvement concentrated in success-level information. Search gains have a standard deviation of 21.4 percentage points, and tested historical estimators explain less than 2\% of this variance in point estimates. Matched-configuration learning curves probe this distinction across training scales. At the reference token price, free selection observations yield a 2.49-point utility opportunity over the best constant action, driven primarily by gain information. Reusable histories, frozen splits, strong baselines and scoring tools provide a common foundation for developing and evaluating \mbox{value-aware search agents.}
\end{abstract}
\ccsdesc[500]{Information systems~Evaluation of retrieval results}
\keywords{Web search agents, marginal value, counterfactual evaluation, probability calibration, adaptive retrieval, execution experience, benchmark}
\begin{teaserfigure}
\centering\includegraphics[width=\textwidth]{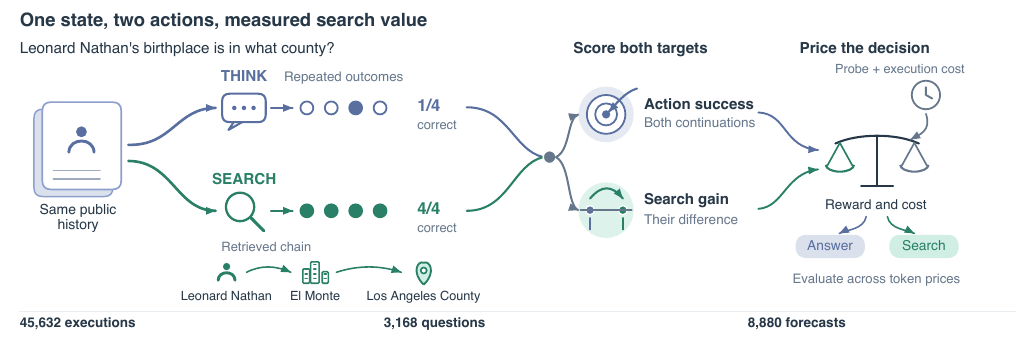}
\caption{One saved question illustrates paired search-value measurement. For Leonard Nathan's birthplace, reference-equivalent answers occur in one of four THINK executions and all four SEARCH executions. Both branches feed success and gain evaluation. The decision stage separately charges probes and continuation costs. The fractions describe \mbox{this selected illustration.}}
\Description{A person-document history forks into thought and search paths. A real entity-city-county chain supports search. Four observed outcomes per branch join at a neutral node and feed both success and gain measurements; a cost scale leads to answer or search.}
\label{fig:teaser}
\end{teaserfigure}
\maketitle
\hypersetup{pdftitle={SearchHorizonBench: A Benchmark for Search Value in AI Agents},pdfauthor={Jingjie Ning, Xueqi Li, Yibo Kong}}

\section{Introduction}
Large-scale agentic search logs document iterative query reformulation and lexical links to earlier retrieved context \citep{wild}. Each additional search can resolve a missing relation or entity while consuming time and money. In Test-400, permitting search increases reference-equivalent answer success from 69.38\% to 73.56\% and mean actor tokens from 1,069 to 6,829. Search-value prediction identifies which questions justify that \mbox{additional information cost.}

SearchHorizonBench defines \emph{search value} as the expected reward difference between continuing with search and answering from the same public history. The actor completes a chosen continuation; the forecaster estimates its outcome. Three measurements describe their behavior. Action-success prediction measures correctness under each continuation. Gain prediction measures their difference. Decision utility combines the selected reward with the cost of acquiring its information. Calibration captures question difficulty; paired gain and cost measurements identify the information relevant to \mbox{choosing further search.}

The benchmark connects these measurements through native paired executions. Every saved state supports both continuations, repeated outcomes, fixed reference answers and recorded resource costs. Historical action means and task priors establish strong references. Repeated executions separate expected-gain error from generation noise. Matched-configuration learning curves measure the effect of additional training data. A decision-information factorial independently varies access to query-specific gains and query-specific costs. These controls give search-value estimators a common test of prediction accuracy and \mbox{decision utility.}

Paired evaluation identifies which information calibration improves. On Test-1632, a historical correction reduces Brier from 0.2145 to 0.1797. Task priors account for 88.9\% of that change, and success-level prediction accounts for 99.9\%. Search gains vary substantially across questions, with a standard deviation of 21.4 percentage points. The calibrated historical predictors explain less than 2\% of this variation in point estimates. The evaluated regressors remain near zero gain skill across same-actor, same-backend training sets of 20 to 1,100 questions. These results establish question-specific gain prediction as a distinct \mbox{benchmark target.}

Gain information identifies opportunities for cost-aware search. At a token price of 0.005 per thousand tokens, a sample-informed selector gains 2.49 utility percentage points over the best constant action. Holding cost information fixed and varying gain information retains 2.48 points; varying cost alone gives 0.68 points. A deployable two-answer consistency rule pays for its probes, making signal acquisition part of evaluation. SearchHorizonBench contributes the paired-execution resource, a controlled account of difficulty and gain information, and an evaluation connecting that information to decisions. Its 45,632 executions and 8,880 forecasts cover 3,168 questions across actors, retrieval backends, \mbox{budgets and rewards.}

\section{Related Work}
\paragraph{Retrieval decisions and agent control.}
Self-RAG, FLARE, DRAGIN and Adaptive-RAG develop retrieval triggers or strategy selection \citep{selfrag,flare,dragin,adaptiverag}. Search-R1 and R$^2$-Searcher train search policies with reinforcement learning \citep{searchr1,r2searcher}. Stop-RAG, HALT, TASR and Agentic Abstention study stopping through values, verification, stability or confidence \citep{stoprag,halt,tasr,agenticabstention}. SearchJev scores legal search decisions \citep{searchjev}. SearchHorizonBench supplies aligned executions for measuring future action value and the experience needed \mbox{to estimate it.}

\paragraph{Execution-derived utility.}
Calibration Is Not Control studies intervention value through matched-prefix execution and evaluates learned controllers \citep{calibration}. CoBRA learns tool-use boundaries from paired internal and external trajectories \citep{cobra}. Search-G1 trains intervention-calibrated readouts of retrieval necessity and evidence reliance \citep{searchg1}. Tian et al.\ predict answer quality and changes across observed retrieval iterations \citep{partialutility}, and Wu et al.\ model tool necessity, utility, and affordability \citep{tocall}. Bridge Evidence measures a document's contribution to later agent actions \citep{bridgeevidence}; reader-specific utility research examines how evidence preferences and intervention effects transfer \citep{readerutility}. Matched revision studies separate re-solving, scaffolding and draft content \citep{revision}; repeat-aware QA audits separate source-induced answer changes from within-source variation \citep{answerchurn}. SkillLearnBench evaluates experience-derived agent skills at content, trajectory and task-outcome levels \citep{skilllearnbench}. SearchHorizonBench measures prospective action-value prediction and historical reuse with repeated outcomes and \mbox{matched historical baselines.}

\paragraph{Budgets, configuration selection, and calibration.}
Budget-aware control and active-retrieval evaluation account for information costs \citep{budgetcontrol,deepcontrol,bcas,activebudget}. Evaluation-budget studies allocate questions, trajectories and repeated answers to improve precision \citep{allocation}. DRAG selects retriever--generator configurations \citep{drag}. Pandora's Router and stage-aware pruning price value estimation and context use \citep{pandora,notworth}. Search as Computation Allocation formalizes costly information acquisition \citep{searchcomputation}. THERMOMETER and CALIBFORMER transfer probability calibration across tasks and models \citep{thermometer,calibformer}. Our statistical tools build on treatment-effect estimation, query-performance prediction, calibration and U-statistics \citep{atheyimbens,qpp_limits,betacalibration,hoeffding1948}.

\section{Learning from Agent Executions}
\subsection{A shared state and two executable actions}
A \emph{state} $x$ is the complete public interaction history available to the agent, including tool responses and returned reasoning text. A \emph{checkpoint} is a saved state at a legal action boundary. At each checkpoint, \THINK\ reasons with existing evidence and returns an answer. \SEARCH\ may retrieve further evidence before answering. The actor, action instructions, retrieval interface, and remaining budget define the \mbox{execution being predicted.}

For answer correctness $Y^a\in\{0,1\}$ under action $a$, the prediction targets are
\begin{equation}
 p_a(x)=\mathbb E[Y^a\mid x],\qquad
 \Delta(x)=p_S(x)-p_T(x).
 \label{eq:target}
\end{equation}
The first quantities are action success probabilities. Their difference is the expected gain from further search. This gain includes the complete continuation, including query planning, returned evidence, and \mbox{final answer generation.}

The main actor uses enabled reasoning, an 8,192-token completion budget, and at most three additional searches. Both branches start from the same history. Exact queries and retrieval parameters identify cached responses, so repeated executions measure actor variation under a shared evidence response. The two actor identifiers are \texttt{deepseek-flash} and \texttt{deepseek-v4-pro}; the local cross-family extension uses Qwen3-4B \citep{qwen3}. Pro and Qwen continuations from Flash-generated histories measure transfer conditional \mbox{on those histories.}

A forecaster receives the public history, actor identity, tools, and continuation instructions. It returns success probabilities $q_T$ and $q_S$ before seeing the future outcomes. The main elicitation uses a structured response, disabled reasoning, and a 2,048-token output cap. Malformed forecasts use the fixed pair $(0.5,0.5)$. Transport errors remain incomplete until recovered. Reference answers and future retrieval results belong to \mbox{the evaluation data.}

\begin{figure}[t]
\centering\includegraphics[width=\columnwidth]{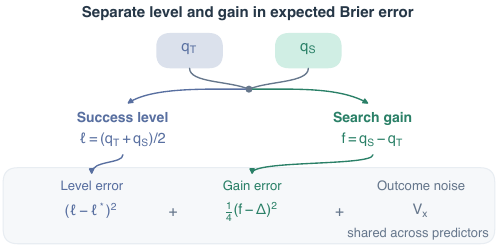}
\caption{One forecast pair supports two targets. Expected Brier error combines success-level error, one quarter of gain error, and shared outcome variance. Paired comparisons cancel the shared variance.}
\Description{The two action probabilities separate into their common success level and action difference. Their squared errors contribute separately to expected Brier error alongside common outcome variance.}
\label{fig:measurement_map}
\end{figure}

\subsection{Interpretable probability corrections}
Four independent executions of each branch provide the development success rates $\bar Y_T$ and $\bar Y_S$. We fit an affine probability correction
\begin{equation}
 \widetilde{\boldsymbol p}(x)=\operatorname{clip}_{[0,1]}
 \{\boldsymbol b+\boldsymbol z(\boldsymbol q(x))B\},
 \qquad \widetilde\Delta=\widetilde p_S-\widetilde p_T.
 \label{eq:correction}
\end{equation}
Here $\boldsymbol z$ standardizes the two raw probabilities using development statistics. Both corrected probabilities use both inputs. Four development-only folds choose a ridge penalty from $\{0.01,0.1,1,10,100\}$. The resulting parameters remain fixed \mbox{during test execution.}

The \emph{historical-mean baseline} predicts each action's development success rate. Task-prior and metadata controls use evaluator-side task categories. Public-prefix ridge uses forecasts and observable history features; text heads use frozen or fine-tuned MiniLM representations. Independent affine calibration uses each action's own probability; shared-input maps use both. Logistic, beta, isotonic, direct-gain and residual models provide additional controls. Two-probe consistency selects a continuation from agreement between independent THINK answers and pays for both probes. Appendix~\ref{app:controls} specifies the features \mbox{and fitting rules.}

Figure~\ref{fig:measurement_map} separates two kinds of predictive information. The \emph{success level} $\ell=(q_T+q_S)/2$ describes how likely the actor is to succeed on the question overall. The \emph{predicted gain} $f=q_S-q_T$ describes the action difference. For the corresponding true level $\ell^*=(p_T+p_S)/2$, average binary Brier error decomposes as
\begin{equation}
 \mathbb E[\overline{\mathrm{Brier}}\mid x]
 = (\ell-\ell^*)^2+\tfrac14(f-\Delta)^2+V_x,
 \label{eq:decomposition}
\end{equation}
where $V_x=\tfrac12\sum_a p_a(1-p_a)$ is the binary-outcome variance for the fixed continuation. Brier is the average squared error of predicted success probabilities against binary correctness. A learned-level/mean-gain control retains question-dependent success levels and uses the historical mean gain. Subtracting this decomposition for two predictors cancels $V_x$ exactly. The resulting paired difference attributes Brier improvement to success-level error and one quarter of gain error. A constant-gain control and an exact zero-gain control give matched tests of \mbox{additional gain information.}

\subsection{Repeated outcomes measure gain error}
Stochastic answers make the difference of two observed success rates noisy. We use independent outcome pairs to estimate the squared expected gain. With $m_a$ executions of each action,
\begin{align}
 \widehat\theta_x={}&
 \frac{\sum_{i\ne j}Y_i^S Y_j^S}{m_S(m_S-1)}
 +\frac{\sum_{i\ne j}Y_i^T Y_j^T}{m_T(m_T-1)}
 -2\bar Y_S\bar Y_T.
 \label{eq:moment}
\end{align}
This U-statistic has conditional expectation $\Delta(x)^2$ \citep{hoeffding1948}. A fixed prediction $f_x$ therefore receives the unbiased gain-risk estimate
\begin{equation}
 \widehat L_x(f)=f_x^2-2f_x(\bar Y_S-\bar Y_T)+\widehat\theta_x.
 \label{eq:risk}
\end{equation}
Paired comparisons use the same executions, and the common final term cancels. We retain finite-sample estimates as computed. The same construction applies to a bounded reward in $[0,1]$, enabling the collection-completion experiment. Appendix~\ref{app:statistics} gives the uncertainty \mbox{and shared-answer calculations.}

\begin{figure*}[t]\centering\includegraphics[width=\textwidth]{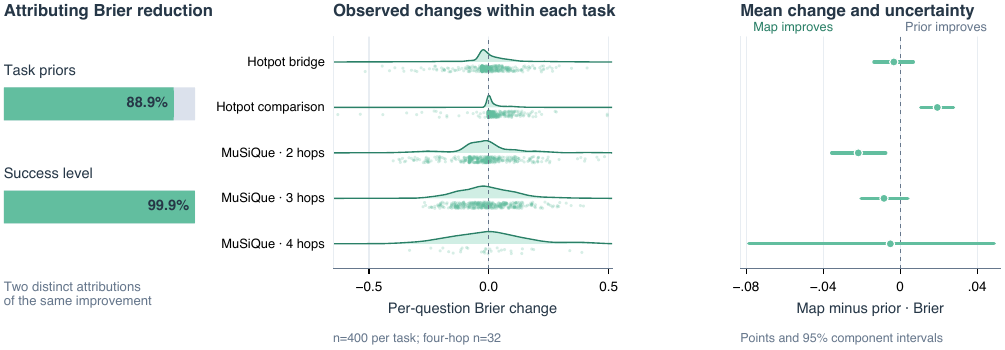}
\caption{Task information and within-task prediction. Left gives two separate attributions of the same Brier reduction. Middle shows observed per-question map-minus-prior Brier changes with density curves. Right shows means and 95\% component intervals. Each stratum has 400 questions except four-hop MuSiQue with 32. Fits use \mbox{development outcomes alone.}}
\Description{Two attribution strips, five per-question Brier-change distributions, and five mean-contrast intervals distinguish task-prior information from within-task prediction.}\label{fig:success_structure}
\end{figure*}

\section{Benchmark and Evaluation}
\subsection{Questions, histories, and transfer settings}
Test-400 contains 120 development and 400 test questions, balanced across HotpotQA bridge and comparison questions and MuSiQue questions requiring two, three, or four reasoning hops \citep{hotpot,musique}. Supporting Wikipedia pages are disjoint between development and test. We save the initial state and states reached after the first and second actual search. Test counts at these checkpoints are 400, 377, and 171. Later checkpoints describe the population of agents that reaches them; repeated-question comparisons \mbox{preserve that dependence.}

Test-1632 contains 1,632 questions with reference-support screening before execution. It includes 400 questions in each of the first four strata and all 32 eligible four-hop questions. Each stratum receives weight 0.2. The frozen source map uses the original 120 Tavily-search questions; target execution uses DeepSeek search. A separate 40/400 development/test study covers Natural Questions queries \citep{naturalquestions}. The benchmark also varies completion budgets of 8,192 or 16,384 tokens and search limits of \mbox{three or six.}

The component study uses 128 development and 320 test questions at a real pending query. Flash and Pro each execute STOP, DeepSeek search, and Serper search. STOP cancels the pending call and answers; the two search actions execute its query and then answer. Four repetitions produce 10,752 executions. A shared STOP execution supports both backend comparisons. ClueWeb22 and FineWeb supply two additional indexed-source conditions on 32 development and 64 test states \citep{clueweb,fineweb,drg}. These comparisons preserve the query and limit each returned packet to five documents with at most \mbox{1,500 characters each.}

A collection study uses 32 development and 96 test tasks from DeepSearchQA \citep{deepsearchqa}. The actor must return a set of requested items. We compare stopping, executing the pending query on the same source, and switching the query to another source. The reward is set F1, the harmonic mean of precision and recall against curated reference items. Four repetitions of three actions yield 1,536 outcomes. The resource contains 45,632 distinct executions and 8,880 forecasts across 3,168 distinct questions; shared executions count once. Appendix~\ref{app:resource} \mbox{provides the inventory.}

\subsection{Scoring and comparison costs}
The common evaluation target is correctness equivalent to the frozen reference answer, scored by the fixed Flash reference-equivalence rule. Normalized exact matches, explicit abstentions and terminal failures receive automatic scores; the remaining distinct question--answer pairs receive semantic judgments. Identical answers share a label. In the actor--retriever audit, two context-isolated AI panels reviewed 812 distinct answers and achieved agreement 95.4\%, with $\kappa=0.918$. Their source-support and answer-equivalence assessments remain separate. A deterministic containment check and exact match evaluate alternative lexical targets in Appendix~\ref{app:reference}.

The stored forecasts and historical corrections are fixed before target execution. Learning curves use separate training and evaluation components within the completed same-configuration bank. Each held-out question receives a prediction from models fitted to other support components. Training subsets, model-selection folds, fitted predictions and original comparison definitions accompany \mbox{the released records.}

Intervals use 10,000 resamples of \emph{support components}, groups of questions linked by a shared supporting page. This grouping preserves source dependence. Equal-stratum analyses normalize the stratum weights within each resample. Reported intervals condition on the fitted predictor. A 0.01 gain-risk difference with the historical paired standard deviation 0.0651 requires about 333 independent questions for 80\% normal-approximation power; Test-400 retains the \mbox{observed source dependence.}

A \emph{router} is a rule that selects a continuation from its forecast. Router utility is $U=Y-\lambda T/1000$, where $T$ includes actor and prediction API tokens and $\lambda$ prices each thousand tokens. We evaluate $\lambda\in\{0,0.001,0.005,0.01,0.02\}$; the focal token valuation is 0.005. If a deployment pays $c$ dollars per thousand total tokens and values an additional correct answer at $v$ dollars, then $\lambda=c/v$. For example, a blended cost of \$1 per million tokens and a \$0.20 value per additional correct answer give $\lambda=0.005$. These are explicit decision-maker valuations; deployment costs depend on the actual token mix. Historical supervision costs, test prediction costs, and shared question-qualification costs are recorded separately. Confidence, entropy, and novelty routers use public inputs and development-selected thresholds. Their feature definitions specify \mbox{the implemented comparisons.}

\begin{figure*}[t]\centering\includegraphics[width=\textwidth]{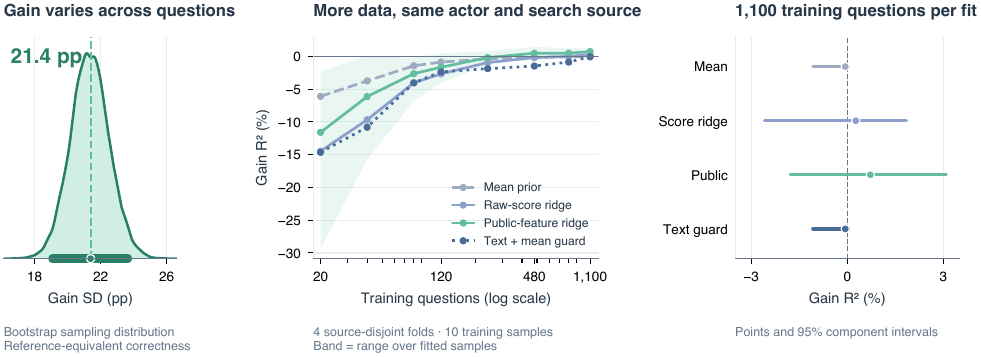}
\caption{Gain prediction under the same actor and search backend. Left gives the sampling distribution of gain standard deviation under reference-equivalent correctness. Middle varies training size using four support-disjoint outer folds and ten nested training samples. The band is the observed range across fitted samples. Right gives 95\% component intervals at 1,100 training questions, conditional on fitted predictors. All model selection uses \mbox{training components alone.}}
\Description{The measured gain variation is contrasted with same-configuration gain-learning curves and their largest-training-set intervals.}\label{fig:gain_learning}
\end{figure*}

\section{Results}
\subsection{Calibration captures task structure}
\begin{table}[t]\centering\small\sffamily\setlength{\tabcolsep}{3.3pt}
\caption{Forecasts on Test-1632 under reference-equivalent correctness. Fits use 120 historical development questions. Public-prefix ridge is the deployable reference; task and metadata priors use evaluator information. Gain $R^2$ is in percent.}\label{tab:forecast_controls}
\begin{tabularx}{\linewidth}{>{\raggedright\arraybackslash}Xrrr}\toprule
\rowcolor{SHBlueLight}Estimator & Brier & Gain MSE & Gain $R^2$\\\midrule
Raw forecast & 0.2374 & 0.0772 & -68.56\\
Global prior & 0.2145 & 0.0460 & -0.50\\
Public-prefix ridge & 0.1796 & 0.0457 & +0.10\\
Calibrated map & 0.1797 & 0.0459 & -0.17\\
Task prior & 0.1836 & 0.0454 & +0.73\\
Metadata ridge & 0.1846 & 0.0451 & +1.51\\
Zero gain + level & 0.1805 & 0.0490 & -7.03\\
\bottomrule\end{tabularx}\end{table}

Execution history provides a strong reference for success prediction. On Test-1632, the frozen map improves Brier from global means' 0.2145 to 0.1797, with paired interval $[-0.0530,-0.0169]$. Task-stratum means score 0.1836 and recover 88.9\% of that observed improvement. The public-prefix ridge baseline scores 0.1796 using observable features. We use it as the deployable reference; stratum and metadata controls explain the contribution of known task structure. Table~\ref{tab:forecast_controls} reports the matched forecasts and \mbox{their gain errors.}

Within-task results specify where finer information helps. On two-hop MuSiQue, the map scores Brier 0.2239 against stratum means' 0.2457, with difference interval $[-0.0356,-0.0075]$. Hotpot comparison questions have success above 95\%, making their 0.0387 task-prior Brier a strong reference; the map scores 0.0579. Figure~\ref{fig:success_structure} gives all five strata and their intervals. The aggregate map-minus-stratum difference is $-0.0039$ with interval $[-0.0197,0.0078]$. These comparisons assign performance credit separately to task-level regularities and \mbox{within-task prediction information.}

The exact decomposition identifies which target improves. Relative to global means, success-level error contributes $-0.034763$ of the map's $-0.034800$ Brier difference, or 99.9\%. The gain term contributes $-0.000037$. The level term further separates into $-0.000293$ from the mean-level shift and $-0.034470$ from variation across questions. Figure~\ref{fig:success_structure} shows the attribution beside the within-task evidence. Together with historical priors, these components identify the information captured by \mbox{each forecast.}

\subsection{Search gains vary widely across questions}
Repeated executions establish a substantial query-specific gain signal. Let $\mu=\mathbb E\Delta(x)$ and $V_\Delta=\mathbb E\Delta(x)^2-\mu^2$. Reference-equivalent correctness gives $\widehat\mu=0.05675$, $\widehat\theta=0.0490$ and gain variance 0.0458, with interval $[0.0364,0.0560]$. Its standard deviation is 21.4 percentage points. This variation defines a question-specific prediction target, evaluated directly through gain error and \mbox{explained variance.}

Gain skill is $R^2_\Delta=1-\widehat L/\widehat V_\Delta$, where $\widehat L$ is generation-noise-corrected gain risk. The historical map achieves $-0.17\%$ with interval $[-1.64,0.87]\%$; public-prefix ridge achieves $+0.10\%$ with interval $[-1.36,1.17]\%$. Table~\ref{tab:forecast_controls} reports all historical estimators. Separate success and gain scores identify the predictive information captured by \mbox{each estimator.}

\subsection{The gain-prediction gap persists with more training}

The same-configuration experiment controls training scale and backend mismatch. It predicts held-out Test-1632 initial states using 20--1,100 training questions from the same Flash actor and DeepSeek search configuration. Raw-score polynomial ridge, public-feature ridge and frozen MiniLM text ridge directly regress the observed gain. Three training-only folds choose regularization; a mean guard also evaluates selecting a constant predictor. Four outer component folds and ten nested samples produce separate fits at \mbox{every training budget.}

Public-feature ridge improves from $-1.67\%$ gain $R^2$ at 120 questions to $+0.72\%$ at 1,100, with final interval $[-1.78,3.09]\%$. Raw-score ridge reaches $+0.27\%$, and the guarded text model selects the mean at the largest budget. Figure~\ref{fig:gain_learning} shows the full curves. The tested families remain near the constant-gain reference across these matched-configuration training budgets. The curves establish a common learning-scale reference for evaluating \mbox{richer decision signals.}

\subsection{Gain information drives price-sensitive opportunity}
\begin{figure*}[t]\centering\includegraphics[width=\textwidth]{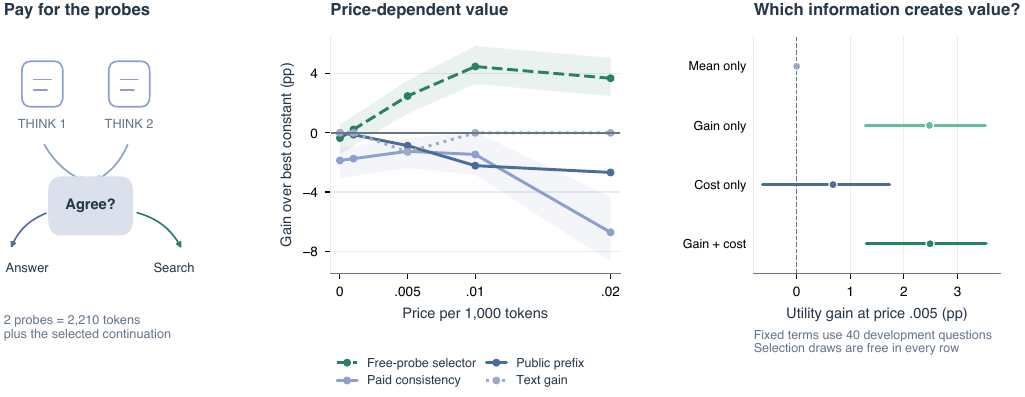}
\caption{Gain information and observation cost. Left illustrates two-probe consistency. Middle compares the five token prices against the best constant, with 95\% component bands. Right varies access to query-specific gain and cost at price 0.005; fixed terms use separate same-configuration development means. Each factorial rule evaluates the same held-out utilities. Its selection draws are free, while consistency pays \mbox{for both THINK probes.}}
\Description{A curved two-probe decision diagram connects to measured price-dependent utility curves and a signed reward/cost contribution display.}\label{fig:decision_prices}
\end{figure*}
The decision factorial separates the information needed for price-sensitive search. Its four rules use fixed or query-specific gain estimates, crossed with fixed or query-specific token estimates. Query-specific values use two selection draws per action; the other two draws evaluate the choice. Fixed values come from forty separate same-configuration development questions. All four rules use the same held-out reward and cost, so their comparison identifies the value of the \mbox{information supplied for selection.}

At $\lambda=0.005$, gain information with fixed costs gives $+2.48$ utility points over the best constant, with interval $[1.28,3.52]$. Cost information with fixed gain gives $+0.68$ points, with interval $[-0.64,1.73]$. Both together give $+2.49$ points; adding cost information to gain information changes utility by $+0.012$ points, with interval $[-0.025,0.051]$. Adding gain information to cost-only selection yields $+1.81$ points, with interval $[1.10,2.65]$. Gain information identifies low-value searches whose cost can be saved. Figure~\ref{fig:decision_prices} keeps the five valuations visible and evaluates \mbox{gain and decision skill separately.}

Disjoint selection and evaluation draws quantify decision opportunity with free selection observations. The selector's four labeled selection executions are excluded from acquisition cost, and its expected value is bounded above by the ideal policy with exact action values. At zero token price, always SEARCH defines the accuracy reference; the selector's difference is $-0.36$ points with interval $[-1.42,0.55]$.

\subsection{Useful decision signals must cover their acquisition cost}
Two-probe consistency selects THINK when two valid normalized THINK answers agree and SEARCH otherwise. Agreement occurs in 65.0\% of probe pairs. Before charging probes, the selected continuation saves 3,451 tokens relative to always SEARCH. The probes consume 2,210, leaving 1,241 net tokens saved at a 1.87-point success cost. Utility is $-1.25$ points at 0.005, with interval $[-2.38,-0.35]$. The benchmark evaluates signal quality together with its \mbox{acquisition cost.}

\subsection{A common benchmark spans execution settings}
The broader resource tests the same quantities under different agent configurations. Indexed retrieval yields Brier 0.1417 on ClueWeb and 0.1498 on FineWeb, against historical means 0.1982 and 0.1981. Qwen actors and forecasters add a separate model family. Budget interventions distinguish evidence effects from generation capacity; increasing the search limit from three to six at 8k tokens raises length failures by 2.5 points, while a 16k budget removes that increase. Collection-completion tasks use bounded set F1 and three actions. Their selected mean-gain predictor and constant STOP policy provide reward-specific references. These settings extend the benchmark's success, gain and decision measurements while retaining each execution \mbox{and reward definition.}

\FloatBarrier
\section{Using the Benchmark}
SearchHorizonBench provides public histories and action definitions for new forecasters. Researchers fit their models on development outcomes, then submit action-success probabilities or signed gain predictions for held-out states. Evaluation uses separate outcome records and supporting-page groups. The value scorer reports success error and generation-noise-corrected gain error; the policy scorer charges prediction and continuation tokens. Saved baselines, fixed priors and reusable splits support \mbox{direct method comparisons.}

Scoring success, gain and utility on matched executions makes a forecaster's contribution to search decisions \mbox{directly testable.}

\section{Conclusion}
SearchHorizonBench establishes a reusable evaluation of when further Web search is worthwhile. Shared histories, repeated continuations and explicit costs connect success prediction, generation-noise-corrected gain estimation and decision utility. The resource spans 45,632 executions. On 1,632 held-out questions, historical calibration improves Brier error by 16.2\%, with task priors accounting for 88.9\% of that change. A search-gain standard deviation of 21.4 percentage points and less than 2\% explained variance in tested historical estimators establish a distinct prediction target. Matched-configuration learning curves provide controlled comparisons across training scales. At the reference token price, factorized decisions identify a 2.49-point free-observation opportunity driven primarily by gain information. The benchmark's public inputs, frozen splits, historical controls and scoring tools make the contribution of execution experience to \mbox{search decisions directly testable.}

\bibliographystyle{ACM-Reference-Format}
\bibliography{references}
\appendix
\section{Data and Execution Settings}\label{app:resource}
\begin{figure*}[t]\centering\includegraphics[width=\textwidth]{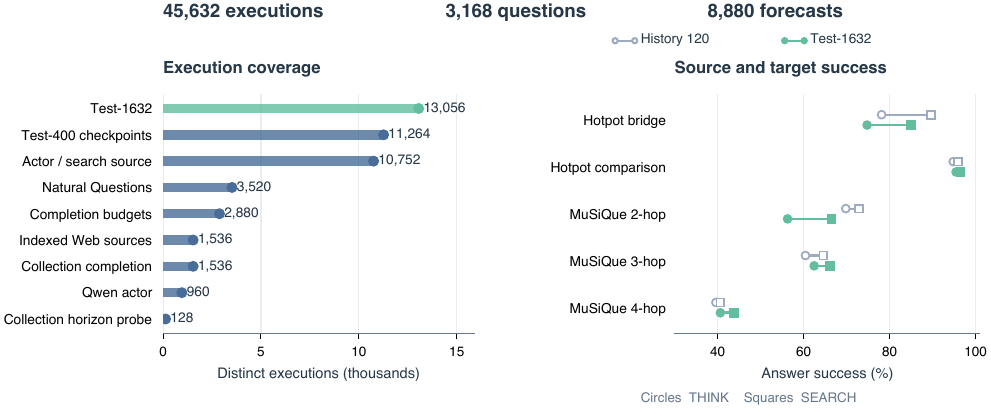}
\caption{Execution coverage and the source/target setting. The nine counts on the left sum to 45,632 unique executions. Reused questions are counted once in the global question total. Right contrasts historical Tavily questions with Test-1632 DeepSeek questions; they are distinct cohorts. Target stratum counts are 400, 400, 400, 400 and 32, with 24 historical questions per stratum. Circles mark THINK and squares SEARCH.}
\Description{Nine execution-count bars and five paired action-success displays show the resource composition and joint question/backend change.}\label{fig:resource_atlas}
\end{figure*}

The resource contains 45,632 distinct executions, 8,880 forecasts and 3,168 questions. Figure~\ref{fig:resource_atlas} counts shared STOP executions once; questions reused across settings retain their identity.

Public predictor records contain histories, actor identity, and action definitions. Evaluation records contain reference answers, observed rewards, support groups, and resource measurements. Saved queries and retrieval responses fix the evidence available to repeated executions. The released value scorer evaluates probability and gain predictions; the policy scorer also charges prediction tokens. Complete prediction matrices, per-case judgments, all pricing settings, and machine-readable manifests \mbox{accompany the resource.}

\paragraph{AI assistance and reproduction.}
DeepSeek supplies executions, forecasts and grading; Qwen supplies cross-family controls. AI also assists auditing, coding, analysis and writing. Scorers and figure sources reproduce the reported comparisons.

\paragraph{Question populations.}
The Test-400 strata contain 24 development and 80 test questions each. Official source splits contribute 160 HotpotQA validation questions, 212 MuSiQue development questions, and 28 MuSiQue training questions to the study test cohort. Its study-development questions and supporting pages are separate. Flash generates the source histories. Later checkpoints and other actors therefore evaluate continuations conditional on Flash reaching those states. The initial-state, first-search and second-search Test-400 populations contain 400, 377 and 171 questions. Their development populations contain 120, 114 and 40. The matched Pro subset contains 80, 74 and 32 \mbox{reached test states.}

Source qualification uses the complete supporting paragraphs. Flash first solves a question with its reference hidden; Flash and Pro then assess reference support separately. Fixed ranks select the eligible cases before actor execution. The Test-1632 campaign screens 8,000 candidates, of which 3,380 receive independent source review and 2,325 are supported. The available four-hop pool yields 32 independently supported questions; all enter the 1,632-question cohort. The five strata receive equal weight, and 1,081 supporting-page components determine resampling. An additional overlap sensitivity removes 149 questions sharing a prior method-selection page and retains 1,483 questions. Its four-hop stratum has one component, so its equal-stratum analysis \mbox{reports point estimates.}

The Natural Questions extension selects 40 development and 400 test queries after source support, standalone clarity and temporal-stability checks. Test pages form 398 components and are development-disjoint. The actor--retriever study selects 128 development and 320 test questions after a valid pending query is reached. Its 848 supporting pages are mutually disjoint and separate from prior labeled studies. This conditional population excludes source trajectories that already answer. The indexed extension samples 32 development and 64 test states by \mbox{fixed family-stratified rank.}

\paragraph{Model and retrieval settings.}
The DeepSeek request identifiers are \texttt{deepseek-flash} and \texttt{deepseek-v4-pro}; records also retain served-model identifiers. Main continuations use enabled reasoning and an 8,192-token total completion budget. The factorial study crosses 8,192/16,384 tokens with three/six permitted searches. Tavily uses basic search with five snippets. Component comparisons normalize DeepSeek and Serper results to five documents and 1,500 characters per document. ClueWeb uses configured shards 0--3 with \texttt{cw22\_a=false}, and FineWeb uses seven shards. Stored responses, timestamps and returned crawl metadata define \mbox{these indexed conditions.}

DeepSeek requests omit temperature, top-$p$ and seed. The provider's documented thinking mode ignores temperature. Its current top-$p$ behavior is model-dependent; Pro thinking fixes top-$p$ at 0.95 with tools and 1.0 otherwise, while Flash non-thinking fixes it at 1.0 \citep{deepseekthinking}. Flash thinking allows top-$p$ in $[0.95,1]$. The historical responses do not echo an effective sampler configuration. We therefore distinguish logged request settings from provider behavior documented on October 25, 2026. Actor requests enable high-effort reasoning; main forecasts disable reasoning with a 2,048-token cap. Repetitions are distinct requests with cached evidence. Conditional generation independence is the statistical assumption; shared identical answers retain \mbox{one correctness judgment.}

\paragraph{Independent model family.}\label{app:qwen_actor}\label{app:qwen_forecaster}
Qwen3-4B Q4\_K\_M uses the pinned revision \texttt{bc640142c66e} and llama.cpp b11669. Actor sampling enables reasoning with temperature 0.6, top-$p$ 0.95, top-$k$ 20, a 16,384-token context and 8,192-token continuation budget. Forecaster sampling disables reasoning with temperature 0.7, top-$p$ 0.8, top-$k$ 20 and a 2,048-token cap. The 120 Qwen forecasts use the same public inputs as the DeepSeek forecasters. The actor extension fixes 240 DeepSeek forecasts and 3,360 fitted probability pairs before Qwen test execution. Its 640 test outputs contain 610 valid answers, nine length failures and 21 malformed outputs; these failure statuses \mbox{receive zero reward.}

\section{Estimators and Inference}\label{app:controls}\label{app:statistics}
\paragraph{Calibration and input controls.}
All fits use development labels. Shared-input maps use both raw probabilities; independent affine maps use one per action. A shared-input control selects a separate penalty for each action. Logistic, beta and isotonic controls provide standard calibration alternatives \citep{betacalibration}. Direct-gain models regress the action difference; residual models adjust a historical map. The release specifies the inputs and fitting grid for each model.

\paragraph{Historical and public-input controls.}
Stratum means use 24 development questions per category. Metadata ridge uses dataset family, comparison membership and hop count. Public-prefix features include both probabilities, log question and tool-text lengths, lexical evidence gap and query/document novelty. The gap is the fraction of question terms absent from tool text. Query novelty is one minus Jaccard overlap with previous queries; document novelty is the fraction of new identifiers. Both are zero before two searches. Four development folds select regularization from $\{0.01,0.1,1,10,100\}$; all controls use the original \mbox{120 development questions.}

\paragraph{Levels and gains.}
The learned-level control predicts the average success probability and assigns a historical mean gain. In the component study, clipping the two reconstructed probabilities to $[0,1]$ can also change their level and intended gain. This occurs on 31 of 320 questions for the reverse Flash mean-gain control. The exact zero-gain control assigns the same learned level to both actions. The own-history mean remains a constant-gain comparator with exactly \mbox{matched development outcomes.}

For a constant historical level $\ell_0$, the level-risk contrast decomposes further into the squared mean-bias difference and the term $\operatorname{Var}(\ell)-2\operatorname{Cov}(\ell,\ell^*)$. The Test-1632 values are $-0.000293$ and $-0.034470$, respectively. The shared variance of observed rewards cancels from this contrast. The latter term includes predictive task differences represented by the stratum controls; the two decompositions identify overall offsets, variation across questions, and action differences \mbox{as separate contributions.}

\paragraph{Gain variance and explained variation.}
For each bootstrap sample, we compute $\widehat V_\Delta=\sum_i w_i\widehat\theta_i-(\sum_iw_i\bar\Delta_i)^2$ and $R^2_\Delta=1-\widehat L/\widehat V_\Delta$ with the same weights. Variance is a plug-in population estimate; the cross-draw second moment removes within-question generation noise. Intervals jointly resample the numerator and denominator and retain negative skill estimates. The reported full-cohort variance denominators remain positive across all bootstrap samples. Each scoring rule defines its own reward and \mbox{gain population.}

\paragraph{Dependence and uncertainty.}
Equations~\ref{eq:moment} and~\ref{eq:risk} use independent generation draws conditional on the cached evidence. Repeated identical answers retain a shared judgment. Supporting-page components define the bootstrap units. Each resample retains all linked questions, actions, checkpoints and repeated labels. Fixed-predictor intervals describe evaluation variation conditional on fitted parameters. The supervision-budget study additionally reports variation over twenty training subsets; its intervals condition \mbox{on those fits.}

\paragraph{Analysis provenance.}\label{app:timing}
The released records retain original comparison definitions, source timestamps, fixed historical predictions and later cross-validation splits. Forecasts used in the held-out execution studies precede their outcomes. The matched-configuration learning curves reuse the completed bank with component-disjoint fitting and evaluation. Their role is measuring training scale under a common execution configuration.

The original sample-size calculation and every comparison criterion accompany the released protocols. Actor-matched family comparisons use a maximum-standardized bootstrap across the declared methods; other displayed intervals describe \mbox{their stated comparisons.}

\paragraph{Text and public-signal models.}\label{app:semantic}\label{app:trainable}\label{app:strata}
Frozen MiniLM sentence embeddings \citep{sbert,minilm} use question text at the initial checkpoint and include the available assistant/tool history at later checkpoints. Ridge and 200-tree forest heads estimate action probabilities or their difference. Fine-tuned heads use sigmoid probability outputs or a tanh gain output, four development folds, encoder learning rates $10^{-5}$ and $3\times10^{-5}$, head rate $10^{-3}$, and epochs $\{1,3,5,10\}$. A mean guard includes the constant action means as a candidate during development selection. Confidence, entropy, query/document novelty and evidence-gap signals use public observations with development-selected thresholds. Evaluator-stratum means use the task category as an explanatory control. Initial-checkpoint Brier is 0.1899 for those means, 0.2035 for the global mean and 0.1836 for \mbox{the shared-input map.}

\section{Answer Scoring and Evidence}\label{app:reference}
\begin{figure*}[t]\centering\includegraphics[width=\textwidth]{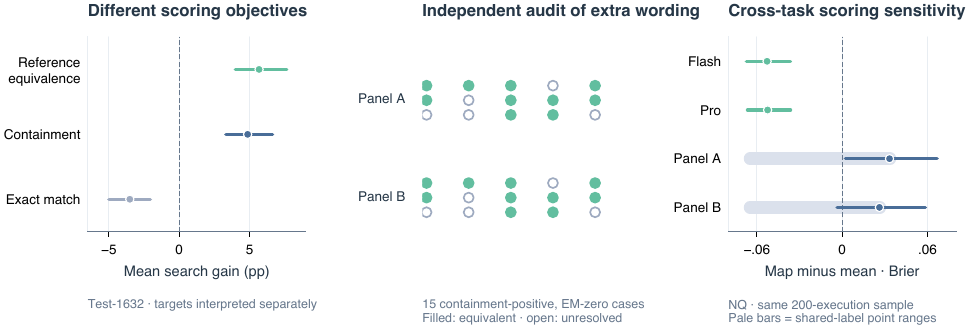}
\caption{Separate scoring objectives and independent audit evidence. Left contrasts the same Test-1632 executions under reference equivalence, normalized containment and exact match. Middle shows the 15 containment-positive, exact-match-zero cases selected from the existing independent AI audit; each panel confirms ten and leaves five unresolved. Right retains the NQ audit comparison on the same 200-execution sample. These objectives and audit projections are \mbox{interpreted separately.}}
\Description{Alternative lexical scoring and independent audit evidence clarify which answer property each score measures.}\label{fig:scoring_targets}
\end{figure*}

\paragraph{Answer scoring.}
Reference-equivalence scoring uses fixed answer strings and semantic judgments. Empty answers, explicit UNKNOWN responses, malformed outputs, generation-length failures and refusals receive zero. Missing transport remains incomplete until recovery. The paired-judge rule keeps disagreements uncertain and records its zero-valued point projection separately from shared-label bounds. An evidence-support audit instead assesses whether the supplied paragraphs establish the requested answer and any additional factual claims. This difference in target explains why a reference-equivalent answer can retain uncertainty \mbox{under source-only evaluation.}

\paragraph{Deterministic containment.}
The containment sensitivity analysis marks an answer correct when a nonempty frozen reference alias occurs as a contiguous token sequence after the benchmark's existing normalization. It removes case, punctuation and articles, normalizes whitespace and assigns zero to failed or UNKNOWN answers. No aliases are added. Test-1632 has mean gain $+0.0486$ with interval $[0.0331,0.0664]$; Test-400 has $+0.0556$ with interval $[0.0287,0.0825]$. Literal case-insensitive containment also gives positive point gains, 0.0505 and 0.0619. These lexical targets measure answer inclusion; the main evaluation uses reference-equivalent correctness. Fifteen inclusion-positive, exact-match-zero cases occur in the existing masked audit. Each independent panel judges ten equivalent and five unresolved, supporting a concrete wording sensitivity without changing the original reward labels.

\paragraph{Independent audits.}
The original 400-question source audit finds 270 supported, 106 insufficient, six ambiguous and eighteen contradicted references. The corresponding question--answer panels retain inverse-inclusion weights and support-component uncertainty. The Natural Questions audit covers 200 sampled executions, representing 178 distinct answers and 155 questions. Its panels agree on 95.5\% of sampled executions, with $\kappa=0.899$, and leave 47 and 44 executions unresolved. Original reward records remain fixed for \mbox{all sensitivity analyses.}

The supported subset of the original 400 questions uses the pre-existing source-only audit and contains 270 questions. Its Brier values are 0.1796 for global means, 0.1637 for stratum means, and 0.1634 for the map. Gain risks are 0.0571, 0.0555 and 0.0576, respectively. Conditioning on source support changes the evaluated population, so these estimates supplement the original full-cohort analysis. Test-1632 uses source-support qualification \mbox{before actor execution.}

The actor-matched audit covers the 83 previously unconfirmed component-study questions, 32 random questions from the 237 confirmed questions, and all other grader-disagreement answers. The union contains 153 questions and 812 distinct answers. Two independent OpenAI/Codex contexts review each batch of at most ten questions. Source-only answers and quotations are sealed before references and anonymized candidates are revealed. The service inherits its default model; an exact served-model snapshot is unavailable. The two panels agree on 775 cases, or 95.4\%, with $\kappa=0.918$. Inverse-inclusion weighting gives agreement 98.3\% and $\kappa=0.965$. The agreement measures describe the \mbox{independent AI panels.}

\paragraph{Shared labels and sample weights.}
A case agreed correct receives interval $[1,1]$, a case agreed incorrect receives $[0,0]$, and each remaining case receives $[0,1]$. The same variable labels every repeated occurrence of that question--answer pair across actions. Flagged-question and original grader-disagreement cases have inclusion probability one. Other answers inherit their question's inclusion probability $32/237$. For candidate gain $f_i$ and comparator $b_i$, the paired contrast is linear in observed rewards,
\begin{equation}
 D_i=f_i^2-b_i^2-2(f_i-b_i)(\bar Y_{i,S}-\bar Y_{i,T}).
\end{equation}
We sum each case's coefficients before varying its uncertain label. Inverse-inclusion-weighted corrections provide point estimates of the lower and upper contrast endpoints. A finite-population variance estimate reports the uncertainty of the sampled corrections separately \mbox{from question-bootstrap variability.}

Finite-population sampling and component resampling retain uncertainty outside the reviewed answers. The actor-matched gain-contrast upper bounds are 0.1583 against the mean-gain comparator and 0.2054 against zero gain. Sealed reviews, sampling probabilities and the complete bound calculations accompany \mbox{the released audit.}

The panels support every factual claim in 188 and 191 answers, compared with 437 and 451 supported target answers. Among 475 sampled answers accepted by both original semantic graders and scored zero by EM, they support 367 and 373 targets and leave 108 and 102 unresolved. These observations distinguish matching the requested answer from validating extra precision. Panel A supports 112 of 153 references and Panel B supports 114; each contradicts two. Separate contexts remove shared conversational history. Both panels use the same reviewer family, so their agreement measures consistency \mbox{within that family.}

For Test-400, allowing all source-flagged answer labels to vary gives adverse intervals $[0.0031,0.0373]$ for gain-minus-raw and $[0.0223,0.0577]$ for Brier-minus-mean. The source-supported subset retains a negative gain-minus-raw interval. The Natural Questions panels yield Brier-minus-source-mean point-projection differences of $+0.0333$ and $+0.0262$, with intervals $[0.0022,0.0667]$ and $[-0.0036,0.0585]$. Their shared-label bounds include both signs. These sensitivities retain the scoring rule and population associated \mbox{with each claim.}

\section{Transfer and Supervision}\label{app:transfer}\label{app:maintenance}
Source corrections reuse earlier executions on target forecasts. Test-1632 fixes the map before execution. Component controls compare same-actor history, direct prediction reuse, historical means and three-pair composition on \mbox{fixed target outcomes.}

Figure~\ref{fig:resource_atlas} compares distinct question cohorts under a joint question/backend change. The component study isolates backend effects through paired executions on \mbox{the same questions.}

\paragraph{Component composition.}
Write A/X for actor A and source X. Backend changes reuse identical STOP outcomes; actor changes acquire the new actor's STOP outcomes. Three-pair history observes A/X, A/Y and B/X using 2,560 distinct outcomes over 128 questions and composes their gains,
\begin{align}
 \widetilde p_{BY,T}&=\widetilde p_{BX,T},\\
 \widetilde\Delta_{BY}&=\widetilde\Delta_{BX}+\widetilde\Delta_{AY}-\widetilde\Delta_{AX},\\
 \widetilde p_{BY,S}&=\operatorname{clip}_{[0,1]}(\widetilde p_{BY,T}+\widetilde\Delta_{BY}).
\end{align}
Components use their own historical labels. A single-pair map costs 1,024 outcomes. Reverse-Flash acquisition, including outputs, reference judgments and calibration forecasts, costs 1,590,189 tokens for same-actor history and 4,712,303 for composition, using 128 and 256 cached search packets. Shared qualification, trajectories and source audits are separately recorded. Test prediction uses one versus two joint requests, averaging 1,073.4 \mbox{and 2,160.0 tokens.}

Under reference equivalence, the actor-matched map-minus-mean gain contrast is $-0.0173$; under exact match it is $+0.0113$. Both corresponding intervals exclude zero. The independent audit's shared-label bounds retain both signs. These results describe how the specified answer objective affects measured gain and remain separate from the history-acquisition costs.

\paragraph{Target STOP feedback.}\label{app:thinkonly}
Four disjoint calibration banks contain sixteen HotpotQA and sixteen MuSiQue questions each. Nested budgets contain 8, 16 or 32 target questions. The STOP offset is the mean difference between observed STOP success and the historical STOP prediction. Signed-opportunity scaling first sets $p_T^\prime=\operatorname{clip}(\widehat p_T+c)$, where $c$ is the observed STOP residual mean. It multiplies positive gains by $(1-p_T^\prime)/\max(1-\widehat p_T,0.05)$ and negative gains by $p_T^\prime/\max(\widehat p_T,0.05)$. Paired controls use both action labels on the same questions and include offset shrinkage, residual affine fits, newly fitted affine maps and means. The shrinkage grid is $\{0,.01,.1,.5,.9,1\}$. The fixed STOP16 rule uses 46.9\% of paired16 incremental tokens and has Brier/gain differences $+0.0036$ and $+0.0020$ against paired offset. Its Brier difference from the fixed source map is $+0.0153$.

\paragraph{Indexed sources and model families.}
ClueWeb/FineWeb calibration uses 32 local development states and 64 test states while retaining the original shared STOP draws. The new SEARCH executions contain 1,518 valid outputs, sixteen length failures and two malformed outputs. The matched local mean gives gain MSE 0.0141/0.0169 for ClueWeb/FineWeb and local affine gives 0.0167/0.0162. Their Brier differences are $-0.0565$ and $-0.0483$, with intervals $[-0.0872,-0.0271]$ and $[-0.0703,-0.0274]$. The released Qwen-actor results retain both forecasters and all scoring rules. The independent Qwen forecaster's gain risk is 0.0322, compared with raw 0.2294 and \mbox{historical mean 0.0315.}

\paragraph{Same-configuration gain learning.}
The additional curves use four saved support-component outer folds, ten nested training samples and budgets 20, 40, 80, 120, 240, 480, 800 and 1,100 questions. The actor, retrieval source and continuation budget remain fixed. Raw-probability polynomial features, public-prefix features and frozen MiniLM vectors feed gain ridge regressors. Three component-grouped training folds choose a penalty from $\{0.1,1,10,100,1000\}$. Weighted fitting gives equal task-stratum mass. A training-only guard additionally compares each family with a constant-gain fit. Scoring averages squared losses over separate fits. Parameter choices, training IDs and \mbox{predictions are released.}

\begin{figure*}[t]\centering\includegraphics[width=\textwidth]{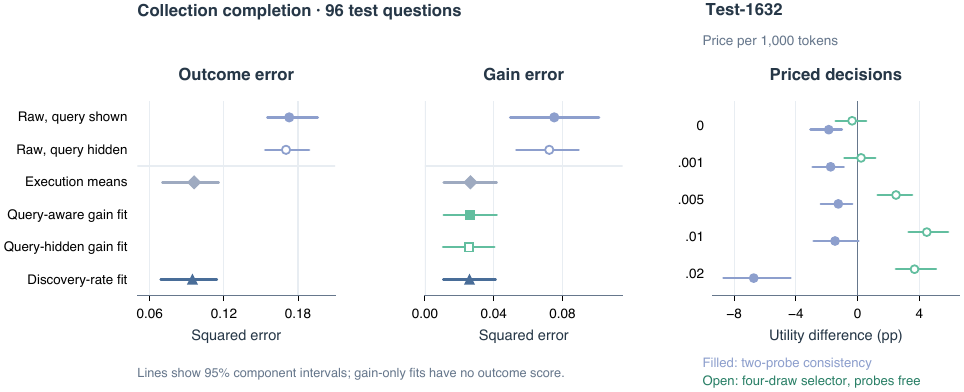}
\caption{Reward and decision controls. The first two panels compare outcome error and two-action-average gain error on 96 held-out collection tasks, with 95\% component intervals. Gain-only fits have no outcome score. The right panel evaluates the separate 1,632-question cohort across all five prices, contrasting paid two-probe consistency with the free-selection-draw reference. These two experiments retain \mbox{their distinct reward definitions.}}
\Description{Collection prediction errors and priced decision intervals show the consequences of changing the reward and paying for decision signals.}\label{fig:collection_audit}
\end{figure*}

\paragraph{Supervision subsets.}\label{app:supervision}
Historical reuse is evaluated with four outer supporting-component folds, three inner folds, twenty training subsets, and target budgets 20, 40, 80, 120 and 240. Losses average separate fitted models, retaining each fit's label budget. At forty questions, the historical120 minus target40 Brier difference is $-0.0110$ with interval $[-0.0142,-0.0083]$. SEARCH alone gives $-0.0103$ with interval $[-0.0137,-0.0071]$. Decomposing squared loss across training subsets attributes 0.0083 of the 0.0110 gap to training-subset variability. The complete curves, cost frontiers and matched-size history controls accompany \mbox{the scoring data.}

\section{Decision and Collection Rewards}\label{app:policy}\label{app:collection}

Two-probe consistency compares two valid normalized THINK answers, selecting THINK on agreement and SEARCH otherwise. UNKNOWN and failed outputs count as disagreement. The remaining two THINK and two SEARCH draws evaluate the selected action. All 36 paired splits are averaged. Cost includes both probes and one selected continuation; CPU time is \mbox{recorded separately.}

The sample-informed selector uses two labeled draws per action and evaluates the chosen action on the remaining pair. Selection executions and judgments are free. For $u_a(x)=\mathbb E[Y^a-\lambda T^a/1000\mid x]$, any randomized selector satisfies $\mathbb E[u_{\pi(x,Z)}(x)]\leq\mathbb E[\max_a u_a(x)]$. Its sampled value is a lower-bound diagnostic for ideal free-probe value in expectation. Deployed rules pay for \mbox{their information acquisition.}

Token valuations express an application tradeoff. Monetary conversion divides the blended token charge by the value of an additional correct answer; actual billing depends on input, output and cache use \citep{deepseekpricing}. For effective throughput $r$ and reward penalty $\beta$ per second, $\lambda=1000\beta/r$. At 100 tokens/second, $\lambda=0.005$ values ten seconds at \mbox{0.5 reward percentage points.}

The information factorial crosses query-specific gains and query-specific costs. Fixed action rewards are 0.68125 and 0.725 from the same-configuration forty-question development set; the fixed incremental token cost is 7,007.49375. Every rule evaluates held-out realized utility. At the focal price, adding gains to cost information improves utility by 1.81 points, with interval $[1.10,2.65]$. The interaction contrast and all five price settings accompany the released analysis.

At zero token price, the ideal oracle improvement over the best constant is $H=(\mathbb E|\Delta|-|\mathbb E\Delta|)/2$. Since $|\Delta|\leq1$, it lies between $\max\{0,(\mathbb E\Delta^2-|\mathbb E\Delta|)/2\}$ and $(\sqrt{\mathbb E\Delta^2}-|\mathbb E\Delta|)/2$. The Test-1632 point moments, $\widehat\theta=0.0490$ and mean gain 0.0568, give the plug-in range $[0,0.0823]$. This algebraic range concerns the ideal zero-price oracle; Figure~\ref{fig:decision_prices} reports achieved disjoint-draw utility at each \mbox{specified token price.}

At $\lambda=0.005$, the Test-400 affine policy's utility difference from SEARCH is $-0.0148$, with interval $[-0.0258,-0.0062]$. The Test-1632 text policy has success 69.05\% at 5,143 tokens, against SEARCH's 71.58\% at 7,613 tokens; its utility difference is $-0.0129$ with interval $[-0.0251,-0.0004]$. The complete five-price comparisons accompany \mbox{the released scores.}

Collection rewards use set precision, recall and F1 against curated atomic reference items, with aliases fixed before execution. STOP cancels the pending query; SAME executes it through DeepSeek and SWITCH through Serper. The query and 8k completion budget are shared. Gain loss averages SAME-minus-STOP and \mbox{SWITCH-minus-STOP risks.}

The collection cohort selects 128 tasks from 225 curated candidates that yield 179 valid pending-query states. Observed-URL/shared-entity groups separate development and test. Four repetitions of each action give 1,536 formal outcomes, including 1,279 valid outputs, 244 invalid outputs and thirteen length failures. The final request retains reasoning and tool history, disables further tools and requests structured items. A distinct development probe contributes 128 outcomes under a three-search continuation and remains outside the formal test comparison. Its F1 contrast against one search is $-0.0386$ with interval $[-0.0713,0.0345]$.

The mean guard returns action means when development selection favors them. The selected query-aware fit selects this guard, reducing raw gain risk by 64.9\%. Query-hidden gain and discovery-rate controls remain separate estimators. The discovery control uses document-level item frequencies without reference aliases, following the motivation of enumeration statistics \citep{crowdenum}. An additional sample-informed analysis retains every constant action and all five fixed token prices. The outcome scorer uses squared error on bounded F1 and the gain scorer uses Equation~\ref{eq:risk}.

\FloatBarrier

\end{document}